\documentclass[letterpaper]{article} 
\usepackage[draft]{aaai25}  
\usepackage{times}  
\usepackage{helvet}  
\usepackage{courier}  
\usepackage[hyphens]{url}  
\usepackage{graphicx} 
\usepackage{natbib}  
\usepackage{caption} 
\usepackage{algorithm}
\usepackage{algorithmic}
\usepackage{amsmath}
\usepackage{amsfonts}
\usepackage{svg}

\usepackage{booktabs}
\usepackage{xcolor}
\usepackage{subcaption}

\usepackage{newfloat}
\usepackage{listings}
\DeclareCaptionStyle{ruled}{labelfont=normalfont,labelsep=colon,strut=off} 
\floatstyle{ruled}
\newfloat{listing}{tb}{lst}{}
\floatname{listing}{Listing}
\title{Symmetry From Scratch:\\ Group Equivariance as a Supervised Learning Task}
\author{
    Haozhe Huang\textsuperscript{\rm 1}\textsuperscript{\rm 2}\textsuperscript{\rm 4}, Leo Kaixuan Cheng\textsuperscript{\rm 1}, Kaiwen Chen\textsuperscript{\rm 1}, Alán Aspuru-Guzik\textsuperscript{\rm 1}\textsuperscript{\rm 2}\textsuperscript{\rm 3}\textsuperscript{\rm 4}
}
\affiliations{
    \textsuperscript{\rm 1}University of Toronto\
    \textsuperscript{\rm 2}Department of Computer Science, University of Toronto\\
    \textsuperscript{\rm 3}Department of Chemistry, University of Toronto\\
    \textsuperscript{\rm 4}Vector Institute for Artificial Intelligence
}

\begin{document}

\maketitle
\begin{abstract}
In machine learning datasets with symmetries, the paradigm for backward compatibility with symmetry-breaking has been to relax equivariant architectural constraints, engineering extra weights to differentiate symmetries of interest. However, this process becomes increasingly over-engineered as models are geared towards specific symmetries/asymmetries hardwired of a particular set of equivariant basis functions. In this work, we introduce \emph{symmetry-cloning}, a method for inducing equivariance in machine learning models. We show that general machine learning architectures (i.e., MLPs) can learn symmetries directly as a supervised learning task from group equivariant architectures and retain/break the learned symmetry for downstream tasks. This simple formulation enables machine learning models with group-agnostic architectures to capture the inductive bias of group-equivariant architectures.
\end{abstract}

%

\section{Introduction}
Equivariance has been crucial to the success of machine learning when working with systems that respect symmetry. From translational invariance used in CNNs for image classification to permutational invariance used in GNNs, their success can be attributed to higher sample efficiency and robustness toward distributional shifts. Nevertheless, enforcing equivariance under the correct problem setting is also essential for better performance. For example, real-life tasks and physical systems may have a lower symmetry level than the expressivity of the equivariant model due to noise or external sources; in such contexts, identifying asymmetries becomes essential for correctly generalizing real-life distributions and applying equivariant models can become too restrictive, capping the model's performance and leading to underfitting. Therefore, a new objective is to account for symmetries and leverage the inductive bias therein while preserving the capability to account for symmetry-breaking. 

\vspace{1.5mm}
Many works \cite{elsayed_revisiting_2020,kaba2024symmetrybreakingequivariantneural,wang_approximately_2022} have focused on adapting equivariant models to account for symmetry-breaking through relaxing known equivariant architectures. This work introduces a much simpler method to approach the problem. Through only supervised learning and existing equivariant models, one could maintain the expressivity of a universal approximator, learn the symmetric architecture of the equivariant model, and proceed with any real-life task that may contain symmetry-breaking data samples -- all without having to design intricate architectures that are tailored towards the symmetries of the data distribution. 

We summarize our key contributions as follows:
\begin{itemize}
    \item We provide empirical evidence that universal function approximators can learn symmetries through supervised learning.
    \item We introduce a simple and novel method for modelling symmetric and symmetry-breaking systems.
    \item We perform a preliminary set of experiments over different symmetry groups and model architectures to validate the generality of our claims.
\end{itemize}

\begin{figure}[ht]
    \centering
    \includesvg{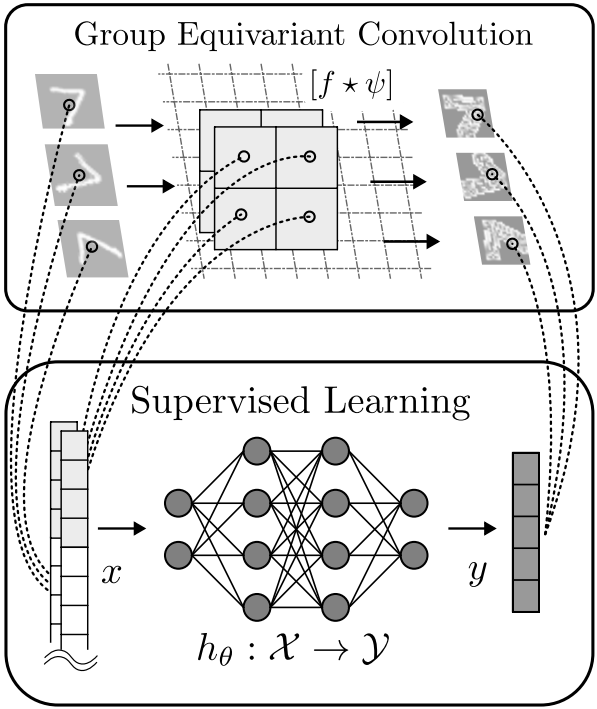}
    \vspace{-9mm}
    \caption{Our model-agnostic symmetry-cloning pipeline for learning approximately equivariant model parameters. The supervised learning framework takes feature maps and the parameters for the equivariant architecture as training data and uses the corresponding output as training labels. The pipeline can be used to train a layer of any group-agnostic architecture.} 
    \label{fig:schematic}
\end{figure}

\subsection{Scope of Work}

We present foundational work on a proof-of-concept training schematic that allows unconstrained, group-agnostic models to learn equivariance directly from equivariant architectures. While this result could shed new insight on topics of model distillation \cite{distill}, model extraction \cite{stealing}, or even further our understanding of neural network training dynamics, in this work, we investigate the efficacy of having learned symmetries as an initial weight condition for group-agnostic models to learn downstream tasks. Additionally, as a proof-of-concept work, we will only focus on feature extraction for images and, therefore, 2D signals over the discrete grid $\mathbb{Z}^2$ using group-equivariant convolutions.

\subsection{Background}
For 2D planar signals, current works that model symmetry-breaking with equivariant models are either fixed MLP layers that comply with relaxed equivariance constraints \cite{kaba2024symmetrybreakingequivariantneural} or constrained to group convolutions \cite{pmlr_v48_cohenc16} with steerable filters \cite{cohen2016steerablecnns} on arbitrarily chosen equivariant bases \cite{wang_approximately_2022}. We show that the types of equivariance enforced by group convolution architectures can actually be learned directly via supervised learning by a more general class of group-agnostic architectures (fig.\ref{fig:schematic}), i.e., general MLPs\footnote{though our methodology easily extends to transformers and other widely used architectures.}, and once trained, can handle tasks involving both symmetric and symmetry-breaking data samples.

\subsubsection{Group Equivariant Convolutions}
Most 2D perception tasks have translational symmetry. Let's consider a discrete linear system for signal processing. The direct consequence of imposing translational equivariance is that any output response of the system is now the convolution of the input signal and its impulse response of the system (i.e., the filter for a convolutional layer). Considering the objective of improved performance for perception tasks, the design decision of layered convolutional neural networks \cite{krizhevsky_imagenet_2012} now only seems natural. 

\vspace{1.5mm}
Building on the success of translational equivariance, group convolutions \cite{pmlr_v48_cohenc16} (GCNNs) were introduced as groundwork for CNNs being, along with translation, equivariant under groups that also included finite rotations and reflections. 
\begin{align}
[f\star \psi](g) &= \sum_{y\in\mathbb{Z}^2}\sum_kf_k(y)\psi_k(g^{-1}y) \label{eq:1}\\
[f\star \psi](g) &= \sum_{h\in G}\sum_kf_k(h)\psi_k(g^{-1}h)\label{eq:2}
\end{align}

Here, the notion of performing inner products with filters under all possible translational transformations is generalized to all possible group transformations. Through a lifting convolution (eq.\ref{eq:1}), we first lift signals from over pixel space to signals over groups (i.e., a semi-direct product of all translations and roto-reflections), then follow up with convolutions performed under (eq.\ref{eq:2}). The critical observation is that while the convolutional operation enforces translational equivariance, roto-reflectional equivariance can be achieved with a stack of transformed filters so that enough information about the transformation is retained over group space \cite{pmlr_v48_cohenc16}.

Vanilla group convolution can only handle the semi-direct product of translation and discrete groups, but with steerable filters \cite{cohen2016steerablecnns,weiler2018learningsteerablefiltersrotation} group convolution can be made to accommodate groups with infinite elements (e.g., the circle group $SO(2)$), effectively expanding CNNs to be equivariant under all isometries $E(2)$ \cite{weiler2021generale2equivariantsteerablecnns}).
 
The feature map for the original group convolution can be seen as coefficients that describe the signal $f: \mathbb{R}^n \times G\rightarrow \mathbb{R}$ with basis chosen such that each axis is associated with a group element. To represent such functions over infinite groups, we leverage that any representation of a compact group $G$ can be written as a direct sum of the group's irreducible representations $\{\rho_j\}_{j=1}^\infty$. Representing $\{\rho_j\}_j$ in a functional form, we can band-limit the signal and thus represent each feature map with finite Fourier coefficients $\{c_i\}$, and recover the entire group representation via feature vector fields $f: \mathbb{R}^n \rightarrow \mathbb{R}^{|c|}$.

\subsubsection{Group Equivariance as a Supervised Learning Task}
While group equivariant convolutions are, in theory, capable of perfect equivariance on planar symmetry groups, when used in practice and planar signals are sampled from a pixel grid $\mathbb{Z}^2$, discretization occurs, equivariance becomes approximate, and complications arise. For example, when designing a steerable basis for $SO(2)$, band-limiting is arbitrary (up to aliasing), and the choice of Gaussian radial profiles is fixed -- not unlikely that a more engineered basis would be more suitable for the learning task. As such, we postulate that there is room for more expressive models to find more optimal group representations while accounting for noise and asymmetries in real-life tasks. 

In our supervised learning formulation, we show that a group-agnostic model can learn the approximate equivariance of group convolutions through only input-output observations and without enforcing any hard constraints on the model's structure or training process, denoting the process as \emph{symmetry-cloning}. Consequently, any symmetries learned by the group-agnostic model from group convolutions could be further optimized end-to-end directly from the task at hand. However, as there is no strong theory behind the convergence of supervised learning, each case of symmetries and class of group-agnostic models must be tested separately. We start this effort by demonstrating symmetry-cloning and its efficacy on downstream tasks with MLPs of different levels of complexity on the translational and discrete rotational group.






\subsection{Related works}
In the more theoretical line of work, tuning higher-symmetry models to lower-symmetry models generally involves some parameter-sharing process. From this perspective, \citet{ravanbakhsh2017equivarianceparametersharing} explored designing model parameters to reflect equivariance over discrete group actions. \citet{shakerinava2024weightsharingregularization} also introduced a weight-sharing regularization scheme, defining a loss that directly encourages weight-sharing to encourage symmetries in parameter space for machine learning in low-data regimes. 

In direct relation to our experimental results, which focus on MLP architectures, is work by \citet{finzi2021practicalmethodconstructingequivariant} on equivariant MLPs (EMLP). They showed that EMLP layers can be constructed by decomposing arbitrary matrix groups (i.e., discrete groups and Lie groups) into their generators. The structure of a single MLP layer under any such group is shown to be the same as the solution of the equivariance constraint over the finite set of generators. While their EMLP architecture is constricted to stacks of EMLP layers with equivariant non-linearities, we show that more general MLP-based architectures, such as the MLP-mixer \cite{tolstikhin_mlp_mixer_2021} (Section \ref{methods}), may also learn group symmetries under a supervised learning context.

Provided that the ground-up approach to equivariant architectures is laborious and computationally expensive, many have worked more broadly on allowing group-agnostic general-purpose architectures to be part of a pipeline that, as a whole and in some limit, can be considered group-equivariant. Notably, frame-averaging introduced by \citet{puny2022frameaveraginginvariantequivariant} and probabilistic symmetrization by \citet{kim2024learningprobabilisticsymmetrizationarchitecture}  both leverage group averaging to convert group-agnostic universal approximators into group-equivariant approximators, while \citet{kaba_equivariance_2023} focused on mapping all samples to their canonical orientation with a learnable canonicalization function. However, for theoretical guarantees, these works often constrain their scope to work within single groups and do not consider symmetry-breaking.

In other works more specific to symmetry breaking, the angle of attack has been engineering modifications or adding additional weights to insert dependence on previously equivariant group transformations. For example, \citet{kaba2024symmetrybreakingequivariantneural} defined a relaxed equivariance constraint that modified the original construction of an EMLP to handle symmetry breaking. For group convolutions, \citet{elsayed_revisiting_2020} tested the practicality of relaxing spatial invariance with a linear combination of a basis set of filter banks; and \citet{wang_approximately_2022} generalized the idea to arbitrary groups with the construction of relaxed group convolution, reintroducing symmetry-breaking dependence on specific pairs of feature map signals and group transformations.

\section{Methods} \label{methods}
\subsection{Symmetry-Cloning}
We propose to use supervised learning for learning group equivariance on a model $h_\theta$ that is not inherently equivariant. Let $\rho: G \rightarrow GL_n(\mathbb{R})$ be a group representation of $G$ and $\Phi_\tau: \mathcal{X} \rightarrow \mathcal{Y}$ be a $G$-equivariant neural network parameterized by $\tau \in \mathcal{T}$ such that $\forall g \in G, x\in\mathcal{X}$:
\begin{equation}
\Phi_\tau(\rho_{in}(g)x) = \rho_{out}(g)\Phi_\tau(x)    
\end{equation}
We train a model $h_{\theta}: \mathcal{X} \rightarrow \mathcal{Y}$ to become approximately equivariant $h_{\theta}(\rho_{in}(g) x) \approx \rho_{out}(g) h_{\theta}(x)$ through 
supervised learning on dataset $\mathcal{D} = \{(x_i, \tau_i), \Phi_{\tau_i}(x_i)\}_i$, where $x_i \sim \mathcal{N}(0,I_{\mathcal{X}})$ and $\tau_i \sim \mathcal{N}(0,I_{\mathcal{T}})$. We call this process \emph{symmetry-cloning} (alg.\ref{alg:1}).

\begin{algorithm}[tb]
\caption{Symmetry-cloning}
\label{alg:1}
\textbf{Input}: $\Phi$ ($G$-equivariant model), $h_\theta$ \\
\textbf{Output}: $G$-cloned $h_\theta$
\begin{algorithmic}[1] 
\WHILE{not converged}
\STATE $x_i \sim \mathcal{N}(0,I_{\mathcal{X}})$ 
\STATE $\tau_i \sim \mathcal{N}(0,I_{\mathcal{T}})$
\STATE $y_i \leftarrow \Phi_{\tau_i}(x_i)$
\STATE \texttt{loss = MSE}$(h_{\theta}(x_i), y_i)$
\STATE \texttt{loss.backpropogate()}
\STATE \texttt{optimizer.update\_weights()}
\ENDWHILE
\STATE \textbf{return} $h_\theta$
\end{algorithmic}
\end{algorithm}
\subsection{Benchmarking Tasks}\label{bt}
\begin{figure}[ht]
    \centering
    \includegraphics[width=0.45\textwidth]{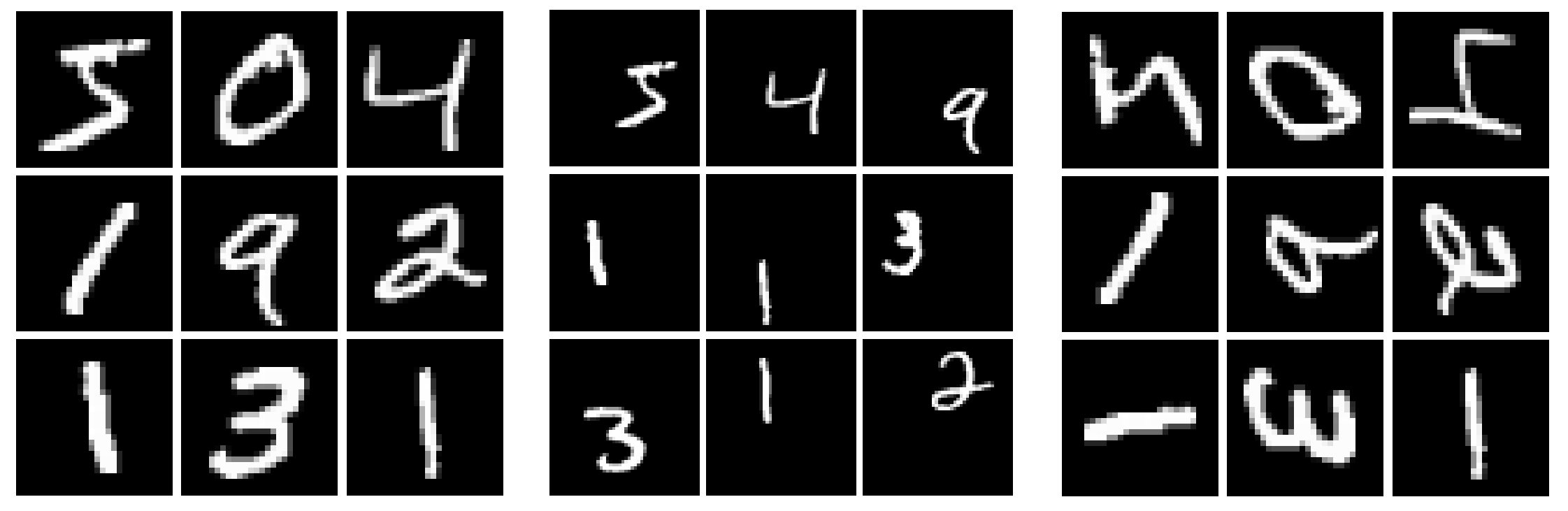}
    \caption{a) original MNIST datset (\emph{left}), b) dataset under $T(2)$ translational tranformations (\emph{middle}), and c) dataset under $C_4$ discrete rotational transformations (\emph{right})}
    \label{fig:dataset}
\end{figure}
To demonstrate the merits of symmetry-cloning, we limit $\rho_{out}$ to be the trivial representation (i.e., group-invariant) for classification tasks and introduce a benchmark consisting of a symmetric task and a symmetry-breaking task. By comparing the performance of group-agnostic models, $G$-cloned models, and group-equivariant models on both tasks, we demonstrate the effectiveness of our method in clonning the symmetries of the groups $T(2)$ and $C_4$, the groups of 2D translations and cyclic rotations by 90 degrees (clockwise), respectively:
\begin{align}
T(n) &:= \{ T_{\mathbf{a}} \ | \ T_{\mathbf{a}}(\mathbf{b}) = \mathbf{a} + \mathbf{b}, \ \mathbf{a}, \mathbf{b} \in \mathbb{R}^n \}, \\
C_n &:= \{ R_{\theta} \ | \ R_{\theta}(\mathbf{x}) = \mathbf{x}R_{\theta}, \ \theta = {2\pi}/{n} \}.
\end{align}

We run the benchmarking tasks for both $T(2)$ and $C_4$ group transformations on the well-studied MNIST handwritten dataset (fig.\ref{fig:dataset}), respectively. In the symmetric task, we test for symmetries baked into the model by evaluating models on transformed samples not encountered by the model during training, while in the symmetry-breaking task, we evaluate the model's ability to differentiate between certain symmetries.

More specifically, we specify the tasks as follows:
\begin{itemize}
    \item\emph{$T(2)$ Symmetric Task:} we train the model on the original dataset (fig.\ref{fig:dataset}a) and evaluate its performance on a test set where every sample is randomly translated within 25\% of the maximum image size (fig.\ref{fig:dataset}b).
    \item\emph{$T(2)$ Symmetry-breaking Task:} we train the model on the translated dataset (fig.\ref{fig:dataset}b) but with labels modified as follows: 
    \[\hat{y} = \begin{cases}
        y &\text{ if shifted left}\\
        y + 10 &\text{ otherwise}
    \end{cases}\] 
    \item\emph{$C_4$ Symmetric Task:} we train the model on the original dataset (fig.\ref{fig:dataset}a) and evaluate its performance on a test set where a random rotation $r\in\{0, \pi/4, \pi/2, 3\pi/4\}$ is randomly applied to each sample (fig.\ref{fig:dataset}c).
    \item\emph{$C_4$ Symmetry-breaking Task:} we train the model on the rotated dataset (fig.\ref{fig:dataset}c) with the labels modified as follows:
    \[\hat{y} = \begin{cases}
        y &\text{ if $r \in \{0, \pi/4\}$}\\
        y + 10 &\text{ otherwise}
    \end{cases}\] 
\end{itemize}

\subsection{Group-agnostic Models}
Building towards increasingly general architectures, we start with a simple case of symmetry-cloning: $T(2)$-cloning a single-channel convolutional layer with a 3x3 kernel and $C_4$-cloning a single-channel group convolutional layer with a 4x3x3 kernel.

\begin{itemize}
    \item\emph{9-block mlp2cnn:} We observe that the convolution operation, when unrolled as a single left matrix multiplication with the image, displays a block Toeplitz pattern. Therefore, the most straightforward group-agnostic architecture would be one that learns the permutation matrix, which, along with kernel parameters, would combine to reconstruct the Toeplitz matrix (fig.\ref{fig:mlpv1}). Note that in this case, the MLP layer is constrained to have as many linear components as there are kernel parameters.
    
    \begin{figure}[ht]
        \centering
        \includesvg{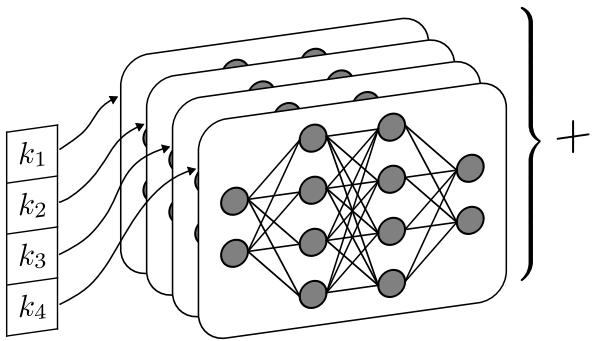}
        \caption{Simple MLP layer component with architecture matching the number of convolution kernel parameters.} 
        \label{fig:mlpv1}
    \end{figure}
    
    \item\emph{approx-mlp2cnn:} We relax the constraint on requiring the architecture to have as many linear components as there are kernels (i.e., having 7, 8, or 10 layers); instead, we add an additional embedding layer followed by a projection to the number of linear components involved. In effect, this allows kernel parameters to act more as input to the group-agnostic model as depicted in (fig.\ref{fig:schematic}).
    
    \item\emph{9-block mlp2gcnn, approx-mlp2gcnn:} In very much the fashion as mlp2cnn, we use four-stacked mlp2cnns to clone a single $C_4$ group-equivariant group convolution layer with a lifting filter of four channels; the approx-mlp2gcnn layers now having four additional projection heads.
\end{itemize}

In both the mlp2cnn and mlp2gcnn cases, we may use the symmetry-cloned MLP layers analogous to how one would stack a full CNN, apply the appropriate pooling, and add a classification head to produce an MNIST classifier. Abusing notation, we also refer to the $T(2)$-cloned classifiers by their layer names. 

Eventually, we wish to apply symmetry-cloning to less constricting architectures on groups with infinite elements. The current procedure for building entire classifiers via stacking symmetry-cloned layers becomes prohibitively expensive as the cloning architectures become more generalized. Nevertheless, without being applied to the benchmarking tasks, we still demonstrate that a more practical architecture, MLP-Mixer, can also be symmetry-cloned.

\begin{itemize}
\item\emph{mlpmixer2cnn:}  We apply a first MLP-Mixer layer to encode the concatenated input and kernel parameters, followed by a second to decode the convolutional output. As the architecture is much more general, large enough MLP-Mixers should be able to learn entire CNNs directly, but we leave that for future work.
\item\emph{mlpmixer2scnn:} Instead of regular group convolutions, we extend symmetry-cloning to a more general class of steerable CNN that is $SO(2)$-equivariant, albeit on a small input size and show that it can roughly capture the equivariant nature of a steerable CNN.
\end{itemize}

\subsection{KL Weight Regularization}
When training symmetry-cloned classifiers on the benchmarking tasks, to better leverage the learned symmetries of symmetry-cloned models, some regularization is needed to ensure that weights do not deviate too quickly from the learned initialization $\theta_{\text{init}}$. Therefore, following \cite{kl}, we use a KL constraint to prevent the weight distribution from drifting away from the equivariant initialization too far too quickly:
\begin{equation}
    \theta_{t+1} \leftarrow \theta_t - \alpha \nabla_{\theta_t} \mathcal{L}(h_{\theta}(x_i), y_i) - \beta \text{KL}(p(\theta_t) || p(\theta_{\text{init}}))
\end{equation}

\section{Results and Discussion}

\begin{figure}[ht]
    \centering
    \includegraphics[scale = 0.2]{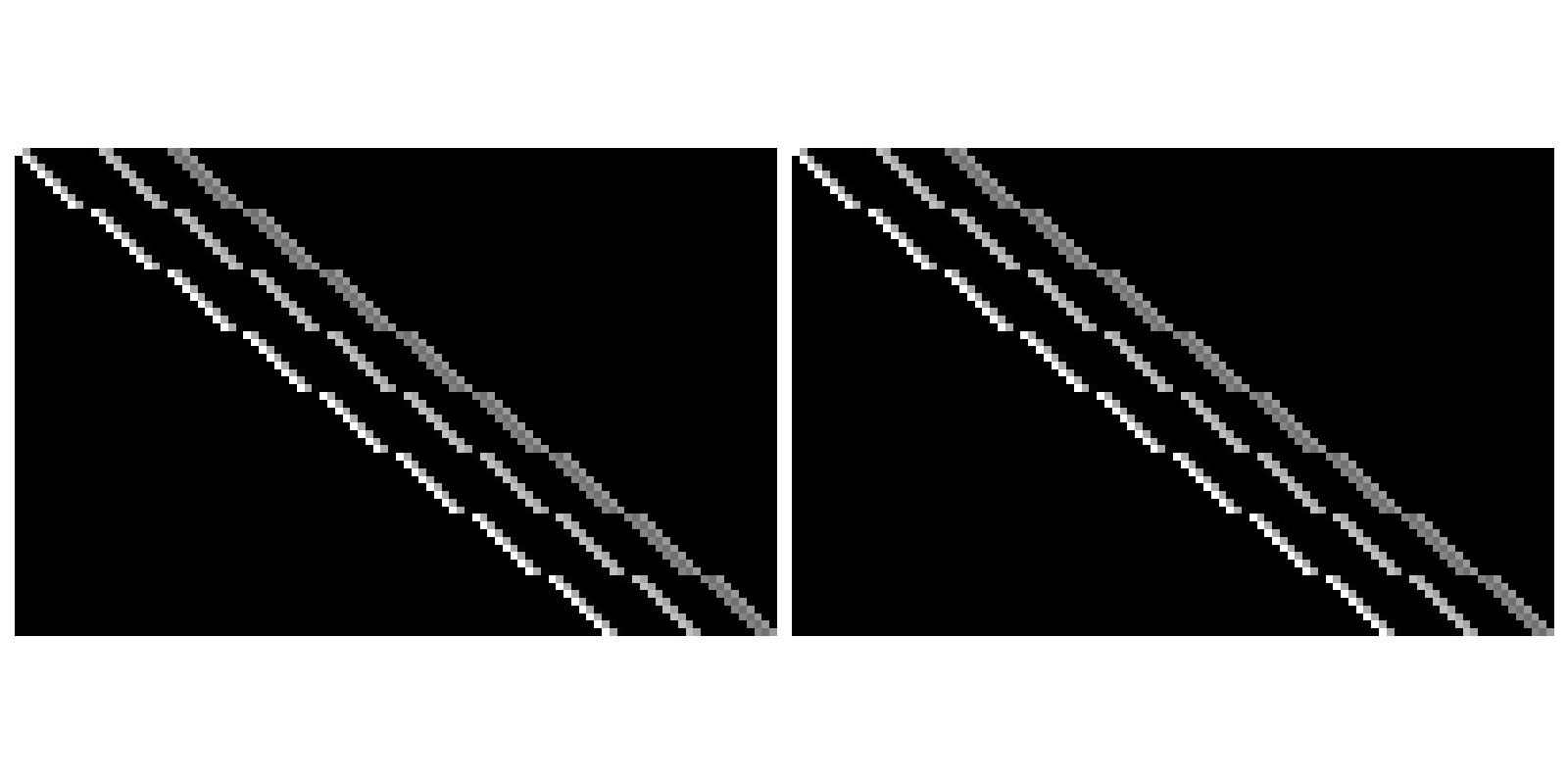}
    \vspace{-7mm}
    \caption{Parameter matrix extracted the learned layer of $T(2)\text{-cloning}$ (\textit{left}); corresponding Toeplitz matrix generated by unrolling convolution (\textit{right}).} 
    \label{fig:Toeplitz_Comparison}
\end{figure}

\begin{figure}[ht]
    \centering
    \includegraphics[scale=0.6]{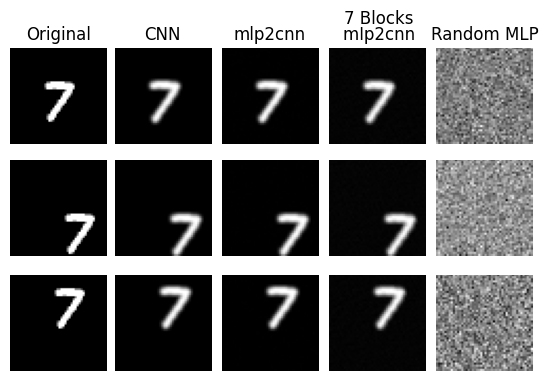}
    \caption{Feature maps with $T(2)$-cloned models. The first column shows the original and translated images, the rest are feature maps from a CNN layer, mlp2cnn, 7-block mlp2cnn, and an untrained MLP layer accordingly.}
    \label{fig:Translational_Featuremap}
\end{figure}

\begin{figure}[ht]
    \centering
    \includegraphics[scale=0.5]{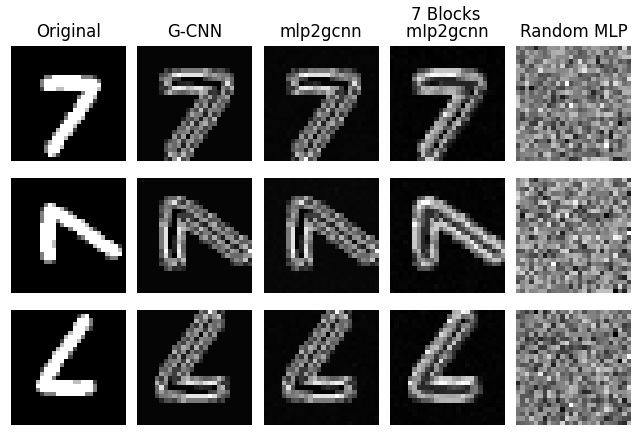}
    \caption{Feature maps with $C_4$-cloned models. The first column shows the original and rotated images, the rest are feature maps from a GCNN layer, mlp2gcnn, 7-block mlp2gcnn, and an untrained MLP layer accordingly.}
    \label{fig:Rotational_Featuremap}
\end{figure}

For symmetry-cloning, in the simplest case of 9-block mlp2cnn, we can compare the learned parameter matrix with the exact Toeplitz matrix unrolled from convolution (fig.\ref{fig:Toeplitz_Comparison}). However, as the cloning model architectures become less constrained with added layers and non-linearity, it becomes harder to unroll and compare with convolutions. So, instead, we show the equivariance of the cloned models via feature map comparisons. For mlp2cnn (fig.\ref{fig:Translational_Featuremap}) and mlp2gcnn models (fig.\ref{fig:Rotational_Featuremap}), symmetry-cloning works exceptionally well, even when the exact Toeplitz correspondence is broken. The performance drop is barely noticeable, and the output feature maps visibly maintain equivariance. This dramatically contrasts with the feature mapping of an MLP layer that has not gone through symmetry cloning.

\begin{figure}[ht]
    \centering
    \includegraphics[scale=0.6]{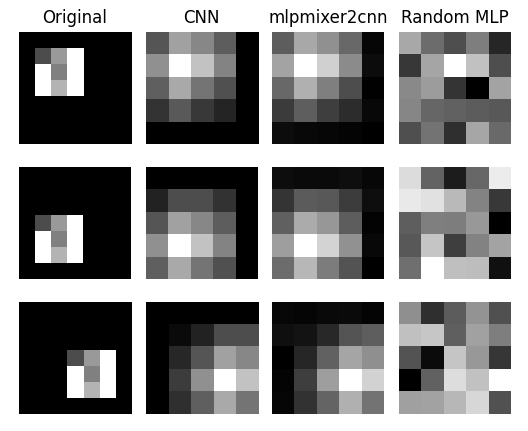}
    \caption{Feature maps with $T(2)$-cloned MLP-Mixer. The first column shows the original and translated images, the rest are feature maps from a CNN layer, mlpmixer2cnn, and an untrained MLP layer accordingly.}
    \label{fig:MixerCNN_Featuremap}
\end{figure}

\begin{figure}[ht]
    \centering
    \includegraphics[scale=0.6]{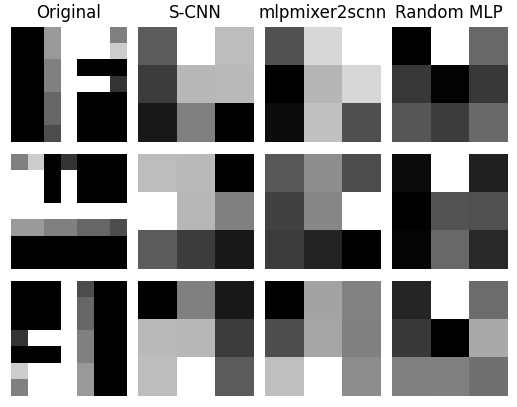}
    \caption{Feature maps with $SO(2)$-cloned MLP-Mixer. The first column shows the original and rotated images, the rest are feature maps from a steerable-CNN layer, mlpmixer2scnn, and an untrained MLP layer accordingly.}
    \label{fig:Mixer_Featuremap}
\end{figure}

We notice a significant increase in computation time required for symmetry-cloning a mlpmixer model. As it becomes nearly prohibitively expensive to train, we revert to much smaller input sizes. Nevertheless, it can be done (fig.\ref{fig:MixerCNN_Featuremap}). Training $SO(2)$-cloned mlpmixer2scnn, the issue is more pronounced, but we can still see that the feature maps still resemble the feature maps of the steerable CNN, exhibiting signs of equivariance (fig.\ref{fig:Mixer_Featuremap}).

We also present results from all benchmarking tasks for mlp2cnn and mlp2gcnn compared to their target CNN/GCNN architectures and uncloned MLP counterparts (table.\ref{table:1}). Freeze denotes that we freeze all mlp2cnn/gcnn layers, training only the kernel parameters that are input to the mlp2cnn/gcnn layers, effectively utilizing the mlp2cnn/gcnn layer as an approximate CNN/GCNN layer. Unfreeze denotes that we allow all weights to be trainable while applying the KL regularization term. From the table, we conclude that preliminary experimental results demonstrate that symmetry-cloned mlp2cnn/gcnn models can learn both symmetric and symmetry-breaking downstream tasks. Although performance improvement over MLP in the rotational symmetry task is not ideal, we have not performed exhaustive hyperparameter searches or architectural optimization for this case.

With group equivariant convolutions as a starting point both figuratively and literally, symmetry-cloning could be framed as an effective weight initializer, warming up the models with an infinite amount of data, ``warm-starting" an architecture with a better initialization. Symmetry-cloning allows models to adapt to a lower symmetry than the one they were trained on. Practically, we show that through simple supervised learning, we can learn a universal function approximator to capture both symmetric and symmetry-breaking features in a dataset without hardcoded feature engineering.

\newcommand{\ra}[1]{\renewcommand{\arraystretch}{#1}}
\renewcommand{\toprule}{\specialrule{1.5pt}{0em}{0em}}
\renewcommand{\bottomrule}{\specialrule{1.5pt}{0em}{0em}}
\begin{table*}\centering
\ra{1.3}
\begin{tabular}{@{}l  rrr lrrr@{}}\toprule
& \multicolumn{2}{c}{Translated MNIST}  & \phantom{abc}& &\multicolumn{2}{c}{Rotational MNIST} \\ 
\cmidrule{2-3} \cmidrule{6-7}
Model & Symmetry  & Symmetry Breaking &\phantom{abc} & Model & Symmetry & Symmetry Breaking\\
\toprule
MLP & $14.71\pm 0.42$ & $\mathbf{97.53\pm0.08}$ & & MLP & $41.06\pm 0.12$ & $\mathbf{97.18\pm0.04}$\\
CNN & $\underline{94.10\pm0.34}$ & $69.52\pm1.37$ & & GCNN & $\mathbf{96.21\pm0.14}$ & $49.04\pm0.12$\\
 mlp2cnn freeze & & && mlp2gcnn freeze & \\
{\footnotesize $\quad 9\ \textit{Blocks}$} & $\mathbf{94.37\pm0.18}$ & -& & {\footnotesize $\quad 9\ \textit{Blocks}$} & $\underline{95.82\pm0.15}$ & -\\
{\footnotesize $\quad 8\ \textit{Blocks}$} & $87.20\pm0.13$& -&&{\footnotesize $\quad 8\ \textit{Blocks}$} & $58.89\pm0.04$ & - \\
{\footnotesize $\quad 7 \ \textit{Blocks}$} & $79.47\pm0.21$& -& &{\footnotesize $\quad 7\ \textit{Blocks}$} & $53.98\pm0.01$ & -\\
 mlp2cnn unfreeze & $77.83\pm0.24$& $\underline{97.33\pm0.06}$&& mlp2gcnn unfreeze & $57.30\pm0.13$ & $\underline{96.70\pm0.20}$\\

\bottomrule
\end{tabular}
\caption{The evaluated accuracy of baseline models and mlp2cnn models on benchmarking tasks described in Section.\ref{bt}. The best accuracy achieved in each task is in bold, and the second best is underlined. We note that all runs were trained on a single A100 GPU with variance calculated from running each experiment 5 times.}\label{table:1}
\end{table*}



\subsection{Limitations}
The work empirically presents most results, and as such, there are no theoretical bounds on the approximate clonability of any symmetry with any unconstrained model. Most tunable hyperparameters were chosen arbitrarily to demonstrate our approach's general applicability. However, this also leaves much room for improvement on finding more rigorous approaches and optimal architectures.

\section{Future Work}
We present a unique learning task at the intersection of equivariant learning, model distillation, and model extraction. Therefore, finding novel connections to more established fields may allow us to study the efficacy of symmetry-cloning in more practical applications. As immediate next steps to this work, we aim to optimize the symmetry-cloning process, explore ways to speed up the convergence process and make the pipeline applicable to a more extensive range of networks. For example, extending symmetry-cloning to $SO(3)$ could allow for more optimized learning of molecules in voxel representations using existing 3D-Unet architectures \cite{unet3d}. 

We also lay the groundwork for a wide range of follow-up research. Including but not limited to extending symmetry-cloning on more complex groups with other exciting architectures, the likes of Transformers \cite{transformer} and Kolmogorov-Arnold Networks \cite{kans}. Additionally, studying the theoretical aspects of symmetry-cloning, such as convergence properties, data requirements, or even the applicability of symmetry-cloning as a metric for approximate equivariance, could be invaluable. We believe these directions could further validate and broaden the applicability of symmetry-cloning in various machine-learning tasks.

\section{Conclusion}
In this work, we offer a novel perspective on equivariant architectures and provide a straightforward method to study the effects of equivariance under a broader spectrum of model architectures. We show that with symmetry-cloning, group-agnostic models can still leverage the inductive biases of equivariant models while retaining capabilities to adapt to symmetry-breaking tasks. In particular, we have shown empirically that general MLP architectures can learn group equivariance from group-convolution models directly through supervised learning. 

\section{Acknowledgements}
The authors would like to acknowledge useful discussions with Luca Thiede and Abdulrahman Aldossary. Resources used in preparing this research were provided by the Digital Research Alliance of Canada. A.A.-G. thanks Anders G. Frøseth for his generous support. A.A.-G. also acknowledges the generous support of Natural Resources Canada and the Canada 150 Research Chairs program.

\bibliography{aaai25}

\begin{thebibliography}{21}
\providecommand{\natexlab}[1]{#1}

\bibitem[{Cohen and Welling(2016{\natexlab{a}})}]{pmlr_v48_cohenc16}
Cohen, T.; and Welling, M. 2016{\natexlab{a}}.
\newblock Group Equivariant Convolutional Networks.
\newblock In Balcan, M.~F.; and Weinberger, K.~Q., eds., \emph{Proceedings of The 33rd International Conference on Machine Learning}, volume~48 of \emph{Proceedings of Machine Learning Research}, 2990--2999. New York, New York, USA: PMLR.

\bibitem[{Cohen and Welling(2016{\natexlab{b}})}]{cohen2016steerablecnns}
Cohen, T.~S.; and Welling, M. 2016{\natexlab{b}}.
\newblock Steerable CNNs.
\newblock arXiv:1612.08498.

\bibitem[{Elsayed et~al.(2020)Elsayed, Ramachandran, Shlens, and Kornblith}]{elsayed_revisiting_2020}
Elsayed, G.~F.; Ramachandran, P.; Shlens, J.; and Kornblith, S. 2020.
\newblock Revisiting spatial invariance with low-rank local connectivity.
\newblock In \emph{Proceedings of the 37th {International} {Conference} on {Machine} {Learning}}, volume 119 of \emph{{ICML}'20}, 2868--2879. JMLR.org.

\bibitem[{Finzi, Welling, and Wilson(2021)}]{finzi2021practicalmethodconstructingequivariant}
Finzi, M.; Welling, M.; and Wilson, A.~G. 2021.
\newblock A Practical Method for Constructing Equivariant Multilayer Perceptrons for Arbitrary Matrix Groups.
\newblock arXiv:2104.09459.

\bibitem[{Hinton, Vinyals, and Dean(2015)}]{distill}
Hinton, G.; Vinyals, O.; and Dean, J. 2015.
\newblock Distilling the Knowledge in a Neural Network.
\newblock arXiv:1503.02531.

\bibitem[{Jaques et~al.(2017)Jaques, Gu, Bahdanau, Hernández-Lobato, Turner, and Eck}]{kl}
Jaques, N.; Gu, S.; Bahdanau, D.; Hernández-Lobato, J.~M.; Turner, R.~E.; and Eck, D. 2017.
\newblock Sequence Tutor: Conservative Fine-Tuning of Sequence Generation Models with KL-control.
\newblock arXiv:1611.02796.

\bibitem[{Kaba et~al.(2023)Kaba, Mondal, Zhang, Bengio, and Ravanbakhsh}]{kaba_equivariance_2023}
Kaba, S.-O.; Mondal, A.~K.; Zhang, Y.; Bengio, Y.; and Ravanbakhsh, S. 2023.
\newblock Equivariance with learned canonicalization functions.
\newblock In \emph{Proceedings of the 40th {International} {Conference} on {Machine} {Learning}}, volume 202 of \emph{{ICML}'23}, 15546--15566. Honolulu, Hawaii, USA: JMLR.org.

\bibitem[{Kaba and Ravanbakhsh(2024)}]{kaba2024symmetrybreakingequivariantneural}
Kaba, S.-O.; and Ravanbakhsh, S. 2024.
\newblock Symmetry Breaking and Equivariant Neural Networks.
\newblock arXiv:2312.09016.

\bibitem[{Kim et~al.(2024)Kim, Nguyen, Suleymanzade, An, and Hong}]{kim2024learningprobabilisticsymmetrizationarchitecture}
Kim, J.; Nguyen, T.~D.; Suleymanzade, A.; An, H.; and Hong, S. 2024.
\newblock Learning Probabilistic Symmetrization for Architecture Agnostic Equivariance.
\newblock arXiv:2306.02866.

\bibitem[{Krizhevsky, Sutskever, and Hinton(2012)}]{krizhevsky_imagenet_2012}
Krizhevsky, A.; Sutskever, I.; and Hinton, G.~E. 2012.
\newblock {ImageNet} {Classification} with {Deep} {Convolutional} {Neural} {Networks}.
\newblock In \emph{Advances in {Neural} {Information} {Processing} {Systems}}, volume~25. Curran Associates, Inc.

\bibitem[{Liu et~al.(2024)Liu, Wang, Vaidya, Ruehle, Halverson, Soljačić, Hou, and Tegmark}]{kans}
Liu, Z.; Wang, Y.; Vaidya, S.; Ruehle, F.; Halverson, J.; Soljačić, M.; Hou, T.~Y.; and Tegmark, M. 2024.
\newblock KAN: Kolmogorov-Arnold Networks.
\newblock arXiv:2404.19756.

\bibitem[{Puny et~al.(2022)Puny, Atzmon, Ben-Hamu, Misra, Grover, Smith, and Lipman}]{puny2022frameaveraginginvariantequivariant}
Puny, O.; Atzmon, M.; Ben-Hamu, H.; Misra, I.; Grover, A.; Smith, E.~J.; and Lipman, Y. 2022.
\newblock Frame Averaging for Invariant and Equivariant Network Design.
\newblock arXiv:2110.03336.

\bibitem[{Ravanbakhsh, Schneider, and Poczos(2017)}]{ravanbakhsh2017equivarianceparametersharing}
Ravanbakhsh, S.; Schneider, J.; and Poczos, B. 2017.
\newblock Equivariance Through Parameter-Sharing.
\newblock arXiv:1702.08389.

\bibitem[{Shakerinava et~al.(2024)Shakerinava, Sohrabi, Ravanbakhsh, and Lacoste-Julien}]{shakerinava2024weightsharingregularization}
Shakerinava, M.; Sohrabi, M.; Ravanbakhsh, S.; and Lacoste-Julien, S. 2024.
\newblock Weight-Sharing Regularization.
\newblock arXiv:2311.03096.

\bibitem[{Tolstikhin et~al.(2021)Tolstikhin, Houlsby, Kolesnikov, Beyer, Zhai, Unterthiner, Yung, Steiner, Keysers, Uszkoreit, Lucic, and Dosovitskiy}]{tolstikhin_mlp_mixer_2021}
Tolstikhin, I.~O.; Houlsby, N.; Kolesnikov, A.; Beyer, L.; Zhai, X.; Unterthiner, T.; Yung, J.; Steiner, A.; Keysers, D.; Uszkoreit, J.; Lucic, M.; and Dosovitskiy, A. 2021.
\newblock {MLP}-{Mixer}: {An} all-{MLP} {Architecture} for {Vision}.
\newblock In \emph{Advances in {Neural} {Information} {Processing} {Systems}}, volume~34, 24261--24272. Curran Associates, Inc.

\bibitem[{Tramèr et~al.(2016)Tramèr, Zhang, Juels, Reiter, and Ristenpart}]{stealing}
Tramèr, F.; Zhang, F.; Juels, A.; Reiter, M.~K.; and Ristenpart, T. 2016.
\newblock Stealing Machine Learning Models via Prediction APIs.
\newblock arXiv:1609.02943.

\bibitem[{Vaswani et~al.(2017)Vaswani, Shazeer, Parmar, Uszkoreit, Jones, Gomez, Kaiser, and Polosukhin}]{transformer}
Vaswani, A.; Shazeer, N.; Parmar, N.; Uszkoreit, J.; Jones, L.; Gomez, A.~N.; Kaiser, L.; and Polosukhin, I. 2017.
\newblock Attention is {All} you {Need}.
\newblock In \emph{Advances in {Neural} {Information} {Processing} {Systems}}, volume~30. Curran Associates, Inc.

\bibitem[{Wang, Walters, and Yu(2022)}]{wang_approximately_2022}
Wang, R.; Walters, R.; and Yu, R. 2022.
\newblock Approximately {Equivariant} {Networks} for {Imperfectly} {Symmetric} {Dynamics}.
\newblock In \emph{Proceedings of the 39th {International} {Conference} on {Machine} {Learning}}, 23078--23091. PMLR.
\newblock ISSN: 2640-3498.

\bibitem[{Weiler and Cesa(2021)}]{weiler2021generale2equivariantsteerablecnns}
Weiler, M.; and Cesa, G. 2021.
\newblock General $E(2)$-Equivariant Steerable CNNs.
\newblock arXiv:1911.08251.

\bibitem[{Weiler, Hamprecht, and Storath(2018)}]{weiler2018learningsteerablefiltersrotation}
Weiler, M.; Hamprecht, F.~A.; and Storath, M. 2018.
\newblock Learning Steerable Filters for Rotation Equivariant CNNs.
\newblock arXiv:1711.07289.

\bibitem[{Özgün Çiçek et~al.(2016)Özgün Çiçek, Abdulkadir, Lienkamp, Brox, and Ronneberger}]{unet3d}
Özgün Çiçek; Abdulkadir, A.; Lienkamp, S.~S.; Brox, T.; and Ronneberger, O. 2016.
\newblock 3D U-Net: Learning Dense Volumetric Segmentation from Sparse Annotation.
\newblock arXiv:1606.06650.

\end{thebibliography}

\appendix

\end{document}